\documentclass[11pt,a4paper]{article}
\usepackage[hyperref]{acl2021}
\usepackage{times}
\usepackage{latexsym}

\usepackage{microtype}
\usepackage{booktabs}       
\usepackage{makecell}       
\usepackage{graphicx}       

\aclfinalcopy 
\def\aclpaperid{12} 

\title{BERT Goes Shopping:\\Comparing Distributional Models for Product Representations}

\author{
    Federico Bianchi \\
    Bocconi University \\
    Milano, Italy \\
    \texttt{f.bianchi@unibocconi.it}\thanks{\textbf{ }Federico and Bingqing contributed equally to this research.} \\\And
    Bingqing Yu \\
    Coveo \\
    Montreal, CA \\
    \texttt{cyu2@coveo.com} \\\And
    Jacopo Tagliabue \\
    Coveo AI Labs \\
    New York, United States \\
    \texttt{jtagliabue@coveo.com}\thanks{\textbf{ }Corresponding author.} \\
}

\date{}

\begin{document}
\maketitle
\begin{abstract}
Word embeddings (e.g., word2vec) have been applied successfully to eCommerce products through~\textit{prod2vec}. Inspired by the recent performance improvements on several NLP tasks brought by contextualized embeddings, we propose to transfer BERT-like architectures to eCommerce: our model -- ~\textit{Prod2BERT} -- is trained to generate representations of products through masked session modeling. Through extensive experiments over multiple shops, different tasks, and a range of design choices, we systematically compare the accuracy of~\textit{Prod2BERT} and~\textit{prod2vec} embeddings: while~\textit{Prod2BERT} is found to be superior in several scenarios, we highlight the importance of resources and hyperparameters in the best performing models. Finally, we provide guidelines to practitioners for training embeddings under a variety of computational and data constraints.
\end{abstract}

\section{Introduction}
Distributional semantics~\cite{Landauer1997AST} is built on the assumption that the meaning of a word is given by the contexts in which it appears: word embeddings obtained from co-occurrence patterns through \textit{word2vec}~\cite{Mikolov2013EfficientEO}, proved to be both accurate by themselves in representing lexical meaning, and very useful as components of larger Natural Language Processing (NLP) architectures~\cite{conneau2017word}.
The empirical success and scalability of \textit{word2vec}  gave rise to many domain-specific models~\cite{Ng2017dna2vecCV,Grover2016node2vecSF,10.1145/3139958.3140054}: in eCommerce,~\textit{prod2vec} is trained replacing words in a sentence with product interactions in a shopping session~\cite{Grbovic15}, eventually generating vector representations of the products. The key intuition is the same underlying~\textit{word2vec} -- you can tell a lot about a product by the company it keeps (in shopping sessions). The model enjoyed immediate success in the field and is now essential to NLP and Information Retrieval (IR) use cases in eCommerce~\cite{Vasile16,BianchiSIGIReCom2020}.

As a key improvement over~\textit{word2vec}, the NLP community has recently introduced~\textit{contextualized representations}, in which a word like \textit{play} would have different embeddings depending on the general topic (e.g. a sentence about \textit{theater} vs \textit{soccer}), whereas in \textit{word2vec} the word \textit{play} is going to have only one vector. Transformer-based architectures~\cite{Vaswani2017AttentionIA} in large-scale models - such as BERT~\cite{devlin2019bert} - achieved SOTA results in many tasks~\cite{nozza2020mask,rogers2020primer}. As Transformers are being applied outside of NLP~\cite{Chen2020GenerativePF}, it is natural to ask whether we are missing a fruitful analogy with product representations. It is \textit{a priori} reasonable to think that a pair of sneakers can have different representations depending on the shopping context: is the user interested in buying these shoes because they are running shoes, or because these shoes are made by her favorite brand? 

In \textit{this} work, we explore the adaptation of~\textit{BERT}-like architectures to eCommerce: through extensive experimentation on downstream tasks and empirical benchmarks on typical digital retailers, we discuss advantages and disadvantages of contextualized embeddings when compared to traditional~\textit{prod2vec}. We summarize our main contributions as follows:

\begin{enumerate}
    \item we propose and implement a BERT-based contextualized product embeddings model (hence,~\textbf{Prod2BERT}), which can be trained with online shopper behavioral data and produce product embeddings to be leveraged by downstream tasks;
    
    \item we benchmark Prod2BERT against~\textit{prod2vec} embeddings, showing the potential accuracy gain of contextual representations across different shops and data requirements. By testing on shops that differ for traffic, catalog, and data distribution, we increase our confidence that our findings are indeed applicable to a vast class of typical retailers;
    
    \item we perform extensive experiments by varying hyperparameters, architectures and fine-tuning strategies. We report detailed results from numerous evaluation tasks, and finally provide recommendations on how to best trade off accuracy with training cost;
    
    \item we share our code\footnote{Code available at~\url{https://github.com/vinid/prodb}}, to help practitioners replicate our findings on other shops and improve on our benchmarks.
\end{enumerate}

\subsection{Product Embeddings: an Industry Perspective}
\label{sec:ecommerce_an_industry_perspective}

The eCommerce industry has been steadily growing in recent years: according to~\citet{uscommerce}, 16\% of all retail transactions now occur online; worldwide eCommerce is estimated to turn into a \$4.5 trillion industry in 2021~\cite{ecomworld}. Interest from researchers has been growing at the same pace~\cite{Tsagkias2020ChallengesAR}, stimulated by challenging problems and by the large-scale impact that machine learning systems have in the space~\cite{ecomamazon}. Within the fast adoption of deep learning methods in the field~\cite{10.1145/3394486.3403278,Zhang2020TowardsPA,Yuan2020ParameterEfficientTF}, product representations obtained through~\textit{prod2vec} play a key role in many neural architectures: after training, a product space can be used directly~\cite{Vasile2016MetaProd2VecPE}, as a part of larger systems for recommendation~\cite{10.1145/3383313.3411477}, or in downstream NLP/IR tasks~\cite{Tagliabue2020ShoppingIT}. Combining the size of the market with the past success of NLP models in the space, investigating whether Transformer-based architectures result in superior product representations is both theoretically interesting and practically important. 

Anticipating some of the themes below, it is worth mentioning that our study sits at the intersection of two important trends: on one side, neural models typically show significant improvements at large scale~\cite{Kaplan2020ScalingLF} -- by quantifying expected gains for ``reasonable-sized'' shops, our results are relevant also outside a few public companies~\cite{CoveoSIGIR2021}, and allow for a principled trade-off between accuracy and ethical considerations~\cite{Strubell2019EnergyAP}; on the other side, the rise of multi-tenant players\footnote{As an indication of the market opportunity, in the space of AI-powered search
and recommendations we recently witnessed Algolia~\cite{AlgoliaRound} and Lucidworks raising 100M USD~\cite{LWRound}, Coveo raising 227M CAD~\cite{CoveoRound}, Bloomreach raising 115M USD \cite{Bloomreach}.} makes sophisticated models potentially available to an unprecedented number of shops -- in this regard, we design our methodology to include~\textit{multiple} shops in our benchmarks, and report how training resources and accuracy scale across deployments. For these reasons, we believe our findings will be interesting to a wide range of researchers and practitioners.

\section{Related Work}
\label{sec:related_work}

\textit{Distributional Models}.~\textit{Word2vec}~\cite{Mikolov2013EfficientEO} enjoyed great success in NLP thanks to its computational efficiency, unsupervised nature and accurate semantic content~\cite{levy-etal-2015-improving,AlSaqqa2019TheUO,conneau2017word}.
Recently, models such as BERT~\cite{devlin2019bert} and RoBERTa~\cite{Liu2019RoBERTaAR} shifted much of the community attention to Transformer architectures and their performance~\cite{Talmor2019MultiQAAE,Vilares2020ParsingAP}, while it is increasingly clear that big datasets~\cite{Kaplan2020ScalingLF} and substantial computing resources play a role in the overall accuracy of these architectures; in our experiments, we explicitly address robustness by~\textit{i}) varying model designs, together with other hyperparameters; and~\textit{ii}) test on multiple shops, differing in traffic, industry and product catalog.

\textit{Product Embeddings}.~\textit{Prod2vec} is a straightforward adaptation to eCommerce of~\textit{word2vec}~\cite{Grbovic15}. Product embeddings quickly became a fundamental component for recommendation and personalization systems~\cite{CasellesDupr2018Word2vecAT,CoveoECNLP202}, as well as NLP-based predictions~\cite{BianchiSIGIReCom2020}. To the best of our knowledge, \textit{this} work is the first to explicitly investigate whether Transformer-based architectures deliver higher-quality product representations compared to non-contextual embeddings.~\citet{eschauzier_2020} uses Transformers on cart co-occurrence patterns with the specific goal of basket completion -- while similar in the masking procedure, the breadth of the work and the evaluation methodology is very different: as convincingly 
argued by~\citet{articleshopperintent}, benchmarking models on unrealistic datasets make findings less relevant for practitioners outside of ``Big Tech''. Our work features extensive tests on real-world datasets, which are indeed representative of a large portion of the mid-to-long tail of the market; moreover, we benchmark several fine-tuning strategies from the latest NLP literature (Section~\ref{sec:exp2}), sharing -- together with our code -- important practical lessons for academia and industry peers. The closest work in the literature as far as architecture goes is~\textit{BERT4Rec}~\cite{Sun2019BERT4RecSR}, i.e. a model based on Transformers trained end-to-end for recommendations. The focus of~\textit{this} work is not so much the gains induced by Transformers in sequence modelling, but instead is the quality of the representations obtained through unsupervised pre-training -- while recommendations are important, the breadth of~\textit{prod2vec} literature~\cite{bianchi-etal-2021-query2prod2vec,bianchi-etal-2021-language,Tagliabue2020ShoppingIT} shows the need for a more thorough and general assessment. Our methodology helps uncover a tighter-than-expected gap between the models in downstream tasks, and our industry-specific benchmarks allow us to draw novel conclusions on optimal model design across a variety of scenarios, and to give practitioners actionable insights for deployment.

\section{Prod2BERT}
\subsection{Overview}
The Prod2BERT model is taking inspiration from BERT architecture and aims to learn context-dependent vector representation of products from online session logs. By considering a shopping session as a ``sentence'' and the products shoppers interact with as ``words'', we can transfer masked language modeling (MLM) from NLP to eCommerce. Framing sessions as sentences is a natural modelling choice for several reasons: first, it mimics the successful architecture of \textit{prod2vec}; second, by exploiting BERT bi-directional nature, each prediction of a masked token/product will make use of past and future shopping choices: if a shopping journey is (typically) a progression of intent from exploration to purchase \cite{Harbich2017DiscoveringCJ}, it seems natural that sequential modelling may capture relevant dimensions in the underlying vocabulary/catalog. Once trained, Prod2BERT becomes capable of predicting masked tokens, as well as providing context-specific product embeddings for downstream tasks.

\begin{figure}
    \includegraphics[width=6.3cm]{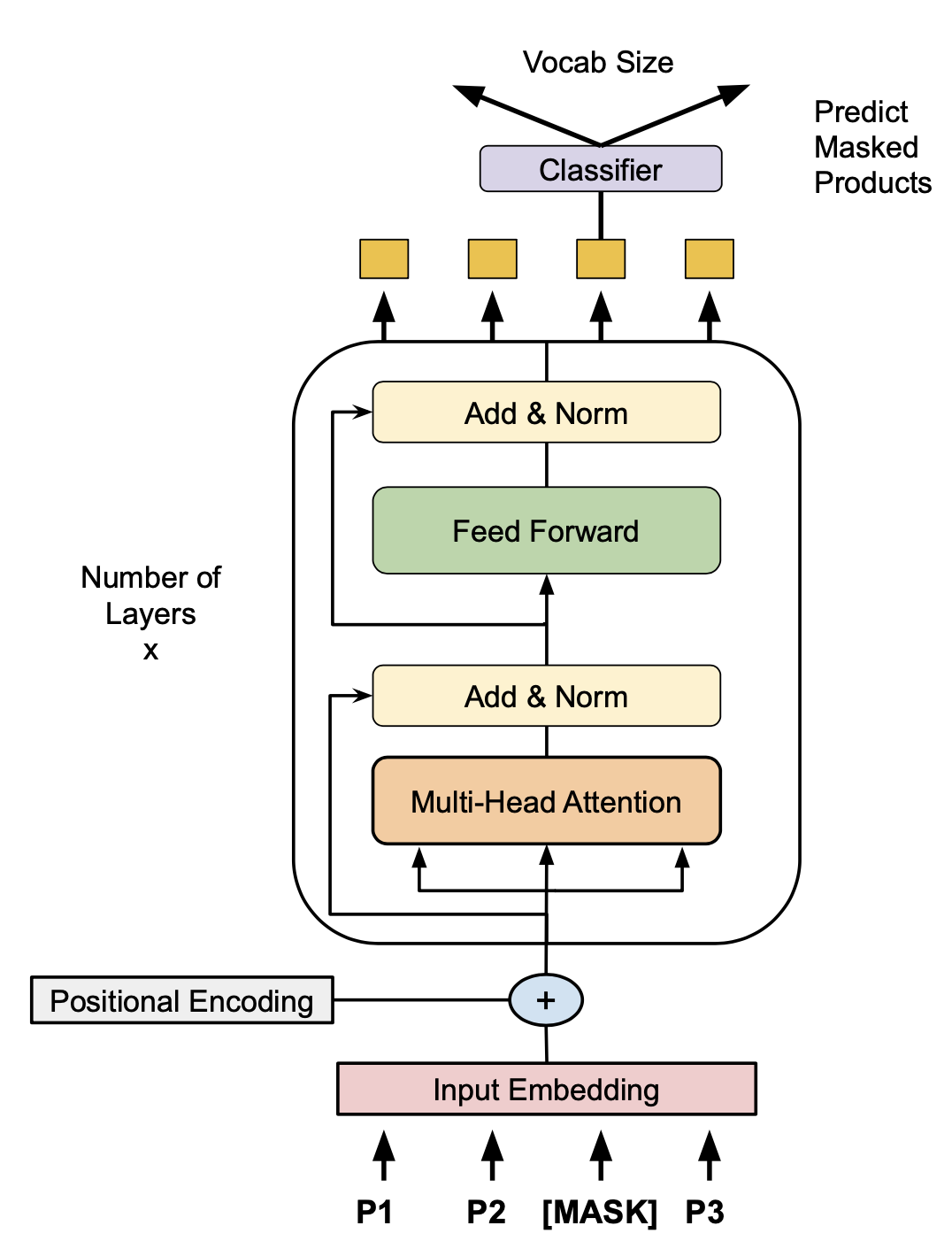} 
    \caption{Overall architecture of Prod2BERT pre-trained on MLM task.}
    \label{fig:emb1}
\end{figure}

\subsection{Model Architecture}
As shown in Figure~\ref{fig:emb1}, Prod2BERT is based on a transformed based architecture~\citet{Vaswani2017AttentionIA}, emulating the successful BERT model. Please note that, different from BERT's original implementation, a white-space tokenizer is first used to split an input session into tokens, each one representing a product ID; tokens are combined with positional encodings via addition and fed into a stack of self-attention layers, where each layer contains a block for multi-head attention, followed by a simple feed forward network. After obtaining the output from the last self-attention layer, the vectors corresponding to the masked tokens pass through a softmax to generate the final predictions.

\subsection{Training Objective}
\begin{figure}
    \includegraphics[width=7.5cm]{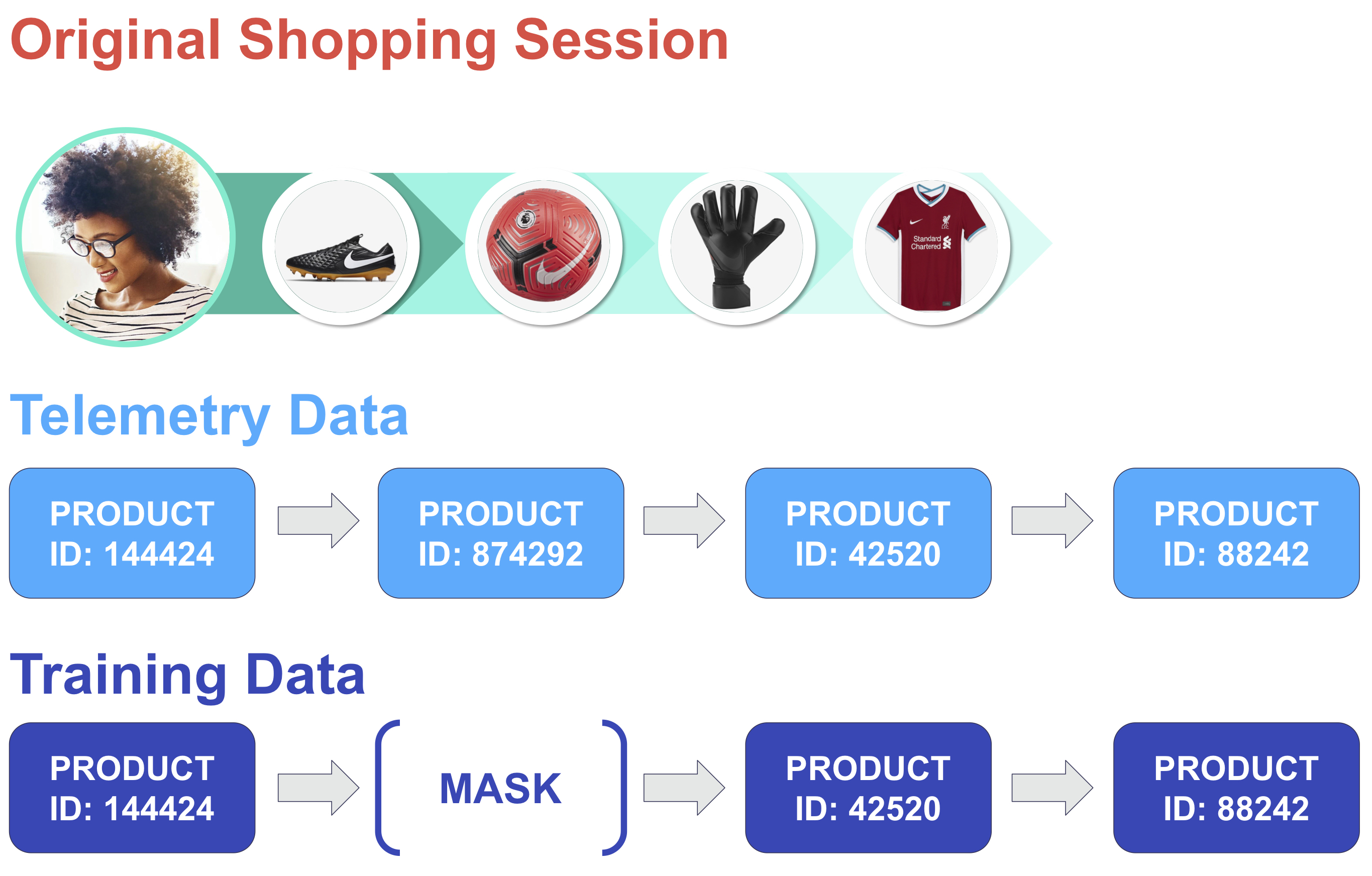} 
    \caption{Transformation of sequential data, from the original data generating process -- i.e. a shopping session --, to telemetry data collected by the SDK, to the masked sequence fed into Prod2BERT.}
    \label{fig:training}
\end{figure}

Similar to~\citet{Liu2019RoBERTaAR,Sun2019BERT4RecSR}, we train Prod2BERT from scratch with the MLM objective. A random portion of the tokens (i.e., the product IDs) in the original sequence are chosen for possible replacements with the~\textit{MASK} token; and the masked version of the sequence is fed into the model as input: Figure~\ref{fig:training} shows qualitatively the relevant data transformations, from the original shopping session, to the telemetry data, to the final masking sequence. The target output sequence is exactly the original sequence without any masking, thus the training objective is to predict the original value of the masked tokens, based on the context provided by their surrounding unmasked tokens. The model learns to minimize categorical cross-entropy loss, taking into account only the predicted masked tokens, i.e. the output of the non-masked tokens is discarded for back-propagation.

\subsection{Hyperparameters and Design Choices}
\label{sec:hyperparams}
There is growing literature investigating how different hyperparameters and architectural choices can affect Transformer-based models. For example,~\citet{Lan2020ALBERTAL} observed diminishing returns when increasing the number of layers after a certain point;~\citet{Liu2019RoBERTaAR} showed improved performance when modifying masking strategy and using duplicated data; finally,~\citet{Kaplan2020ScalingLF} reported slightly different findings from previous studies on factors influencing Transformers performance. Hence, it is worth studying the role of hyperparameters and model designs for Prod2BERT, in order to narrow down which settings are the best given the specific target of our work, i.e.~\textit{product representations}. Table~\ref{tab:hyperdescription} shows the relevant hyperparameter and design variants for Prod2BERT; following improvement in data generalization reported by~\citet{Liu2019RoBERTaAR}, when $duplicated=1$ we augmented the original dataset repeating each session 5 times.\footnote{This procedure ensures that each sequence can be masked in 5 different ways during training.} We set the embedding size to $64$ after preliminary optimizations: as other values offered no improvements, we report results only for one size.

\begin{table}\centering
  \begin{tabular}{ll}
    \toprule
    Parameter & Values\\
    \midrule
    \textit{\# epochs} [\textit{e}] & 10, 20, 50, 100\\
    \textit{\# layers} [\textit{l}] & 4, 8\\
    \textit{masking probability} [\textit{m}] & 0.15, 0.25\\
    \textit{duplicated} [\textit{d}] & 1, 0\\
  \bottomrule
\end{tabular}
  \caption{Hyperparameters and their ranges.}
    \label{tab:hyperdescription}
\end{table}

\section{Methods}

\subsection{Prod2vec: a Baseline Model}
We benchmark Prod2BERT against the industry standard~\textit{prod2vec}~\cite{Grbovic15}. More specifically, we train a CBOW model with negative sampling over shopping sessions~\cite{Mikolov2013EfficientEO}. Since the role of hyperparameters in~\textit{prod2vec} has been extensively studied before~\cite{CasellesDupr2018Word2vecAT}, we prepare embeddings according to the best practices in~\citet{BianchiSIGIReCom2020} and employ the following configuration: $window=15$, $iterations=30$, $ns\_exponent=0.75$, $dimensions=[48, 100]$. While~\textit{prod2vec} is chosen because of our focus on the
quality of the learned representations -- and not just performance on sequential inference \textit{per se} -- it is worth nothing that kNN~\cite{Latifi2020SessionawareRA} over appropriate spaces is also a surprisingly hard baseline to beat in many practical recommendation settings. It is worth mentioning that for both \textit{prod2vec} and Prod2BERT we are mainly interested in producing a dense space capturing the latent similarity between SKUs: other important relationships between products (substitution~\cite{Zuo2020AFL}, hierarchy~\cite{NIPS2017_59dfa2df} etc.) may require different embedding techniques (or extensions, such as interaction-specific embeddings~\cite{10.1145/3366424.3386197}).

\subsection{Dataset}
We collected search logs and detailed shopping sessions from two partnering shops,~\textbf{Shop A} and~\textbf{Shop B}: similarly to the dataset released by~\citet{articleshopperintent}, we employ the standard definition of ``session'' from Google Analytics\footnote{\url{https://support.google.com/analytics/answer/2731565?hl=en}}, with a total of five different product actions tracked:~\textit{detail},~\textit{add},~\textit{purchase},~\textit{remove},~\textit{click}\footnote{Please note that, as in many previous embedding studies~\cite{CasellesDupr2018Word2vecAT,BianchiSIGIReCom2020}, action type is not considered when preparing session for training.}.
Shop A and Shop B are mid-sized digital shops, with revenues between 25 and 100 millions USD/year; however, they differ in many aspects, from traffic, to conversion rate, to catalog structure: Shop A is in the sport apparel category, whereas Shop B is in home improvement. Sessions for training are sampled with undisclosed probability from the period of March-December 2019; testing sessions are a completely disjoint dataset from January 2020. After pre-processing\footnote{We only keep sessions that have between 3 and 20 product interactions, to eliminate unreasonably short sessions and ensure computation efficiency.}, descriptive statistics for the training set for Shop A and Shop B are detailed in Table~\ref{tab:data}. For fairness of comparison, the exact same datasets are used for both Prod2BERT and~\textit{prod2vec}. 

Testing on fine-grained, recent data from~\textit{multiple} shops is important to support the internal validity (i.e. ``is this improvement due to the model or some underlying data quirks?'') and the
external validity (i.e. ``can this method be applied robustly across deployments, e.g.~\citet{10.1145/3383313.3411477}''?) of our findings. 

\begin{table}
\centering
\begin{tabular}{lccc}  \toprule
   Shop & Sessions & Products & 50/75 $pct$\\ \midrule
   \textit{Shop A} & 1,970,832& 38,486 &5, 7\\
   \textit{Shop B} & 3,992,794& 102,942&5, 7\\
   \bottomrule
\end{tabular}
\caption{Descriptive statistics for the training dataset. $pct$ shows $50^{th}$ and $75^{th}$ percentiles of the session length.}
\label{tab:data}
\end{table}

\section{Experiments}
\subsection{Experiment \#1: Next Event Prediction}
\label{sec:experiment_nep}
Next Event Prediction (NEP) is our first evaluation task, since it is a standard way to evaluate the quality of product representations~\cite{letham2013sequential,CasellesDupr2018Word2vecAT}: briefly, NEP consists in predicting the next action the shopper is going to perform given her past actions. Hence, in the case of Prod2BERT, we mask the last item of every session and fit the sequence as input to a pre-trained Prod2BERT model\footnote{Note that this is similar to the word prediction task for cloze sentences in the NLP literature~\cite{petroni2019language}.}. Provided with the model's output sequence, we take the top $K$ most likely values for the masked token, and perform comparison with the true interaction. As for~\textit{prod2vec}, we perform the NEP task by following industry best practices~\cite{BianchiSIGIReCom2020}: given a trained~\textit{prod2vec}, we take all the before-last items in a session to construct a session vector by average pooling, and use kNN to predict the last item\footnote{Previous work using LSTM in NEP~\cite{10.1145/3383313.3411477} showed some improvements over kNN; however, the differences cannot explain the gap we have found between~\textit{prod2vec} and Prod2BERT. Hence, kNN is chosen here for consistency with the relevant literature.}. Following industry standards,~\textit{nDCG@K}~\cite{mitra2018an} with $K=10$ is the chosen metric\footnote{We also tracked~\textit{HR@10}, but given insights were similar, we omitted it for brevity in what follows.}, and all tests ran on $10,000$ testing cases (test set is randomly sampled first, and then shared across Prod2BERT and~\textit{prod2vec} to guarantee a fair comparison).

\subsubsection{Results}
\label{result:sec}

\begin{table}[h!]
\centering
\small
\begin{tabular}{cccc} \toprule
\textit{Model} & Config & \textbf{Shop A} & \textbf{Shop B} \\ \midrule
Prod2BERT & \thead{$ e=10, l=4,$ \\$m=0.25, d=0$ }& 0.433  & 0.259\\\midrule
Prod2BERT&\thead{$ e=5, l=4,$ \\$m=0.25, d=1$ }  & \textbf{0.458} &  \textbf{0.282}\\ \midrule
Prod2BERT&\thead{$ e=10, l=8,$ \\$m=0.25, d=0$ }  & 0.027 &  0.260\\ \midrule
Prod2BERT&\thead{$ e=100, l=4,$ \\$m=0.25, d=0$ }  & 0.427 &  0.255\\ \midrule
Prod2BERT&\thead{$ e=10, l=4,$ \\$m=0.15, d=0$ }  & 0.416 &  0.242 \\ \midrule \midrule
\textit{prod2vec}&$dimension=48$    & \underline{0.326} &  0.214 \\\midrule
\textit{prod2vec}&$dimension=100$   & 0.326 & \underline{0.218} \\  \bottomrule
\end{tabular}
\caption{~\textit{nDCG@10} on NEP task for both shops with Prod2BERT and~\textit{prod2vec} (\textbf{bold} are best scores for Prod2BERT; \underline{underline} are best scores for~\textit{prod2vec}).}
\label{tab:nep}
\end{table}

Table~\ref{tab:nep} reports results on the NEP task by highlighting some key configurations that led to competitive performances. Prod2BERT is significantly superior to~\textit{prod2vec}, scoring up to 40\% higher than the best~\textit{prod2vec} configurations. Since shopping sessions are significantly shorter than sentence lengths in~\citet{devlin2019bert}, we found that changing masking probability from 0.15 (value from standard BERT) to 0.25 consistently improved performance by making the training more effective. As for the number of layers, similar to~\citet{Lan2020ALBERTAL}, we found that adding layers helps only up until a point: with $l = 8$, training Prod2BERT with more layers resulted in a catastrophic drop in model performance for the smaller Shop A; however, the same model trained on the bigger Shop B obtained a small boost. Finally, duplicating training data has been shown to bring consistent improvements: while keeping all other hyperparameters constant, using duplicated data results in an up to 9\% increase in~\textit{nDCG@10}, not to mention that after only 5 training epochs the model outperforms other configurations trained for 10 epochs or more. 

While encouraging, the performance gap between Prod2BERT and~\textit{prod2vec} is consistent with Transformers performance on sequential tasks~\cite{Sun2019BERT4RecSR}. However, as argued in Section~\ref{sec:ecommerce_an_industry_perspective}, product representations are used as input to many downstream systems, making it essential to evaluate how the learned embeddings generalize outside of the pure sequential setting. Our second experiment is therefore designed to test how well contextual representations transfer to other eCommerce tasks, helping us to assess the accuracy/cost trade-off when difference in training resources between the two models is significant: as reported by Table~\ref{tab:costs},
the difference (in USD) between \textit{prod2vec} and Prod2BERT is several order of magnitudes.\footnote{Costs are from official AWS pricing, with 0.10 USD/h for the~\textit{c4.large} (\url{https://aws.amazon.com/it/ec2/pricing/on-demand/}), and 12,24 USD/h for the~\textit{p3.8xlarge} (\url{https://aws.amazon.com/it/ec2/instance-types/p3/}). While obviously cost optimizations are possible, the ``naive'' pricing is a good proxy to appreciate the difference between the two methods.}

\begin{table}
\centering
\begin{tabular}{lcc}  \toprule
   Model & Time A-B & Cost A-B \\ \midrule
   \textit{prod2vec} & 4-20 & 0.006-0.033\$\\
   \textit{Prod2BERT} & 240-1200 & 48.96-244.8\$\\
   \bottomrule
\end{tabular}
\caption{Time (minutes) and cost (USD) for training one model instance, per shop:~\textit{prod2vec} is trained on a~\textit{c4.large} instance, Prod2BERT is trained (10 epochs) on a~\textit{Tesla V100 16GB} GPU from ~\textit{p3.8xlarge} instance.}
\label{tab:costs}
\end{table}

\subsection{Experiment \#2: Intent Prediction}
\label{sec:exp2}
A crucial element in the success of Transformer-based language model is the possibility of adapting the representation learned through pre-training to new tasks: for example, the original~\citet{devlin2019bert} fine-tuned the pre-trained model on 11 downstream NLP tasks. However, the practical significance of these results is still unclear: on one hand,~\citet{Li2020OnTS,Reimers2019SentenceBERTSE} observed that sometimes BERT contextual embeddings can underperform a simple GloVe~\cite{pennington-etal-2014-glove} model; on the other,~\citet{Mosbach2020OnTS} highlights catastrophic forgetting, vanishing gradients and data variance as important factors in practical failures. Hence, given the range of downstream applications and the active debate on transferability in NLP, we investigate how Prod2BERT representations perform when used in the~\textit{intent prediction} task.

\textit{Intent prediction} is the task of guessing whether a shopping session will eventually end in the user adding items to the cart (signaling purchasing intention). Since small increases in conversion can be translated into massive revenue boosting, this task is both a crucial problem in the industry and an active area of research~\cite{Toth2017PredictingSB,articleshopperintent}. To implement the intent prediction task, we randomly sample from our dataset $20,000$ sessions ending with an add-to-cart actions and $20,000$ sessions without add-to-cart, and split the resulting dataset for training, validation and test. Hence, given the list of previous products that a user has interacted with, the goal of the intent model is to predict whether an add-to-cart event will happen or not. We experimented with several adaptation techniques inspired by the most recent NLP literature~\cite{Peters2019ToTO,Li2020OnTS}:

\begin{enumerate}
    \item \textit{Feature extraction (static)}: we extract the contextual representations from a target hidden layer of pre-trained Prod2BERT, and through average pooling, feed them as input to a multi-layer perceptron (MLP) classifier to generate the binary prediction. In addition to alternating between the first hidden layer (\textit{enc\_0}) to the last hidden layer (\textit{enc\_3}), we also tried concatenation (\textit{concat}), i.e. combining embeddings of all hidden layers via concatenation before average pooling.
    \item \textit{Feature extraction (learned)}: we implement a linear weighted combination of all hidden layers (\textit{wal}), with learnable parameters, as input features to the MLP model~\cite{Peters2019ToTO}.
    \item \textit{Fine-tuning}: we take the pre-trained model up until the last hidden layer and add the MLP classifier on top for intent prediction (\textit{fine-tune}). During training, both Prod2BERT and task-specific parameters are trainable.

\end{enumerate}

As for our baseline, i.e.~\textit{prod2vec}, we implement the intent prediction task by encoding each product within a session with its~\textit{prod2vec} embeddings, and feeding them to a LSTM network (so that it can learn sequential information) followed by a binary classifier to obtain the final prediction.

\subsubsection{Results}

\begin{table}
\centering
\begin{tabular}{cccc}  \toprule
   Model&\textit{Method} & Shop & Accuracy \\ \midrule
   Prod2BERT&\textit{enc\_0} & \textit{Shop B} &\textbf{0.567}\\
   Prod2BERT&\textit{enc\_3} & \textit{Shop B} &0.547\\
   Prod2BERT&\textit{concat} & \textit{Shop B} &0.553\\
   Prod2BERT&\textit{wal} & \textit{Shop B} &0.543\\
   Prod2BERT&\textit{fine-tune} & \textit{Shop B} &0.560\\ 
   \textit{prod2vec}& - & \textit{Shop B} &0.558\\\midrule \midrule
   Prod2BERT&\textit{enc\_0} & \textit{Shop A} &0.593\\
   \textit{prod2vec} & - & \textit{Shop A} &\textbf{0.602}\\
   \bottomrule
\end{tabular}
\caption{Accuracy scores in the intent prediction task (best scores for each shop in \textbf{bold}).}
\label{tab:purchase_results}
\end{table}

From our experiments, Table~\ref{tab:purchase_results} highlights the most interesting results obtained from adapting to the new task the best-performing Prod2BERT and~\textit{prod2vec} models from NEP. As a first consideration, the shallowest layer of Prod2BERT for feature extraction outperforms all other layers, and even beats concatenation and weighted average strategies\footnote{This is consistent with~\citet{Peters2019ToTO}, which states that inner layers of a pre-trained BERT encode more transferable features.}. Second, the quality of contextual representations of Prod2BERT is highly dependent on the amount of data used in the pre-training phase. Comparing Table~\ref{tab:nep} with Table~\ref{tab:purchase_results}, even though the model delivers strong results in the NEP task on Shop A, its performance on the intent prediction task is weak, as it remains inferior to~\textit{prod2vec} across all settings. In other words, the limited amount of traffic from Shop A is not enough to let Prod2BERT form high-quality product representations; however, the model can still effectively perform well on the NEP task, especially since the nature of NEP is closely aligned with the pre-training task. 
Third, fine-tuning instability is encountered and has a severe impact on model performance. Since the amount of data available for intent prediction is not nearly as important as the data utilized for pre-training Prod2BERT, overfitting proved to be a challenging aspect throughout our fine-tuning experiments. 
Fourth, by comparing the results of our best method against the model learnt with~\textit{prod2vec} embeddings, we observed~\textit{prod2vec} embeddings can only provide limited values for intent estimation and the LSTM-based model stops to improve very quickly; in contrast, the features provided by Prod2BERT embeddings seem to encode more valuable information, allowing the model to be trained for longer epochs and eventually reaching a higher accuracy score. As a more general consideration -- reinforced by a qualitative visual assessment of clusters in the resulting vector space --, the performance gap is~\textit{very small}, especially considering that long training and extensive optimizations are needed to take advantage of the contextual embeddings.

\section{Conclusion and Future Work}

Inspired by the success of Transformer-based models in NLP,~\textit{this} work explores contextualized product representations as trained through a BERT-inspired neural network,~\textit{Prod2BERT}. By thoroughly benchmarking Prod2BERT against~\textit{prod2vec} in a multi-shop setting, we were able to uncover important insights on the relationship between hyperparameters, adaptation strategies and eCommerce performances on one side, and we could quantify for the first time quality gains across different deployment scenarios, on the other. If we were to sum up our findings for interested practitioners, these are our highlights:

\begin{enumerate}
    \item Generally speaking, our experimental setting proved that pre-training Prod2BERT with Mask Language Modeling can be applied successfully to sequential prediction problems in eCommerce. These results provide independent confirmation for the findings in~\citet{Sun2019BERT4RecSR}, where BERT was used for in-session recommendations over academic datasets. However, the tighter gap on downstream tasks suggests that Transformers' ability to model long-range dependencies may be more important than pure representational quality in the NEP task, as also confirmed by human inspection of the product spaces (see Appendix~\ref{app:visualization_session_embedding} for comparative t-SNE plots).
    \item Our investigation on adapting pre-trained contextual embeddings for downstream tasks featured several strategies in feature extraction and fine-tuning. Our analysis showed that feature-based adaptation leads to the peak performance, as compared to its fine-tuning counterpart.
    \item Dataset size~\textit{does} indeed matter: as evident from the performance difference in Table~\ref{tab:purchase_results}, Prod2BERT shows bigger gains with the largest amount of training data available. Considering the amount of resources needed to train and optimize Prod2BERT (Section~\ref{result:sec}), the gains of contextualized embedding may not be worth the investment for shops outside the top 5k in the Alexa ranking\footnote{See \url{https://www.alexa.com/topsites}.}; on the other hand, our results demonstrate that with careful optimization, shops with a large user base and significant resources may achieve superior results with Prod2BERT.
\end{enumerate}

While our findings are encouraging, there are still many interesting questions to tackle when pushing Prod2BERT further. In particular, our results require a more detailed discussion with respect to the success of BERT for textual representations, with a focus on the differences between words and products: for example, an important aspect of BERT is the tokenizer, that splits words into subwords; this component is absent in our setting because there exists no straightforward concept of ``sub-product'' -- while far from conclusive, it should be noted that our preliminary experiments using categories as ``morphemes'' that attach to product identifiers did not produce significant improvements.
We leave the answer to these questions -- as well as the possibility of adapting Prod2BERT to even more tasks -- to the next iteration of this project. 

As a parting note, we would like to emphasize that Prod2BERT has been so far the largest and (economically) more significant experiment run by~\textit{Coveo}: while we~\textit{do} believe that the methodology and findings here presented have significant practical value for the community, we also recognize that, for example, not all possible ablation studies were performed in the present work. As \newcite{bianchi2021gap} describe, replicating and comparing some models is rapidly becoming prohibitive in term of costs for both companies and universities. 
Even if the debate on the social impact of large-scale models often feels very complex~\cite{Thompson2020TheCL,10.1145/3442188.3445922} -- and, sometimes, removed from our day-to-day duties -- Prod2BERT gave us a glimpse of what unequal access to resources may mean in more meaningful contexts. While we (as in ``humanity we'') try to find a solution, we (as in ``authors we'') may find temporary solace knowing that good ol'~\textit{prod2vec} is still pretty competitive.

\section{Ethical Considerations}
User data has been collected by~\textit{Coveo} in the process of providing business services: data is collected and processed in an anonymized fashion, in compliance with existing legislation. In particular, the target dataset uses only anonymous uuids to label events and, as such, it does not contain any information that can be linked to physical entities.

\bibliographystyle{acl_natbib}
\bibliography{acl2021}

\appendix

\section{Visualization of Session Embeddings}
\label{app:visualization_session_embedding}

\begin{figure}[!b]\centering
    \includegraphics[width=6.3cm]{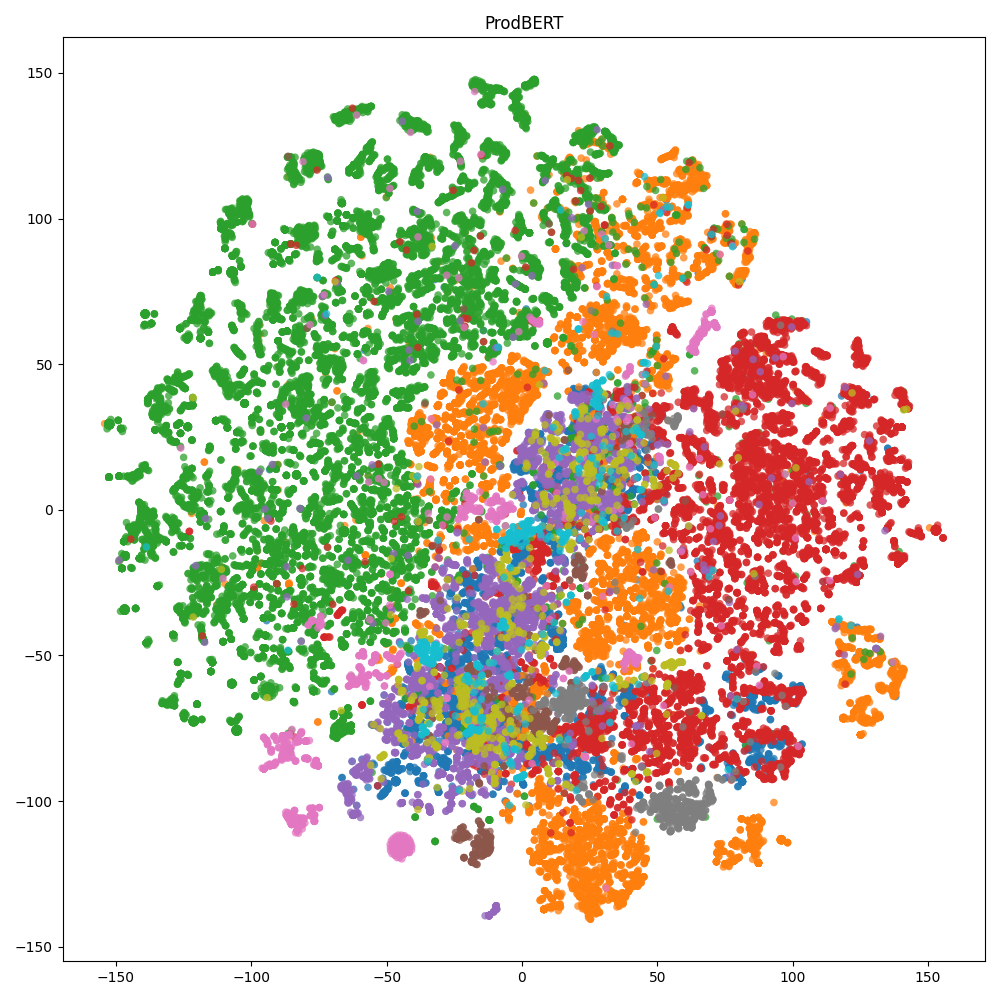} 
    \caption{T-SNE plot of browsing session vector space from Shop A and built with the first hidden layer of pre-trained Prod2BERT.}
    \label{fig:tsne_ms_pb}
\end{figure}

\begin{figure}[!htb]\centering
    \includegraphics[width=6.3cm]{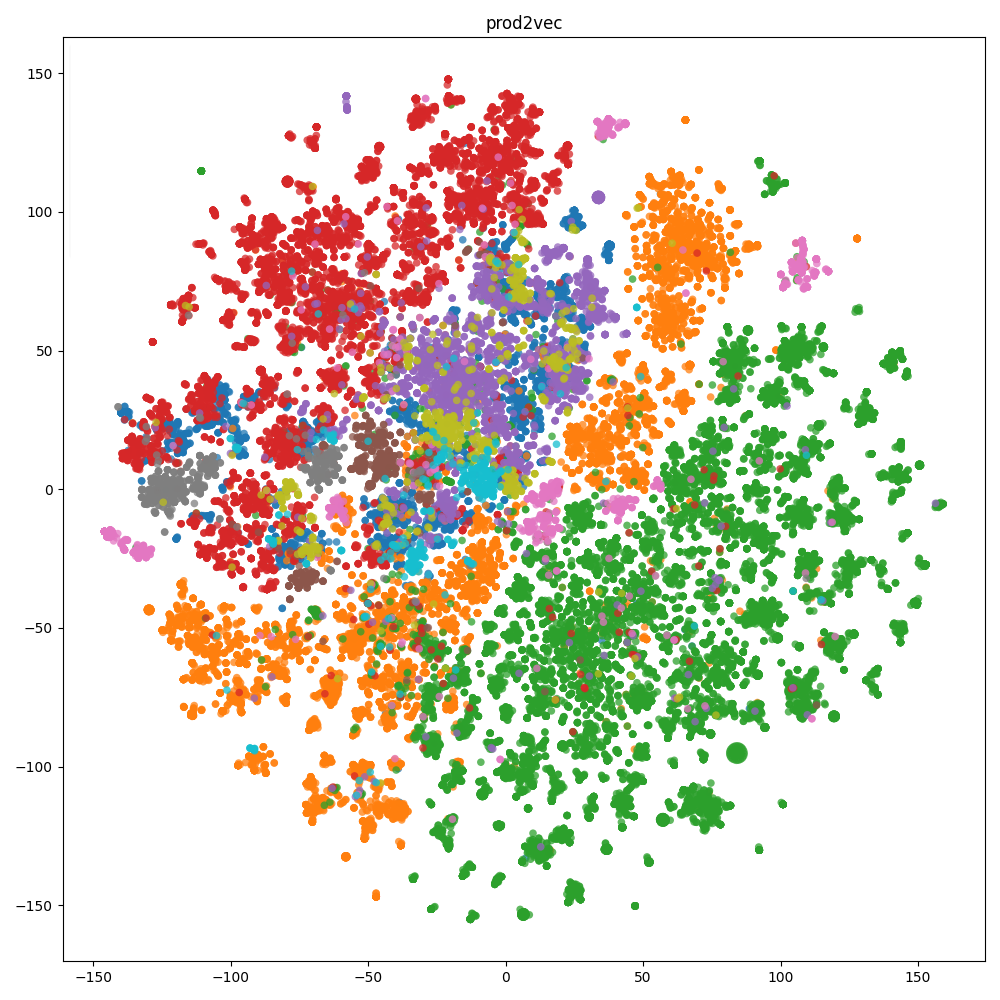} 
    \caption{T-SNE plot of browsing session vector space from Shop A and built with~\textit{prod2vec} embeddings.}
    \label{fig:tsne_ms_p2v}
\end{figure}

\begin{figure}[!htb]\centering
    \includegraphics[width=6.3cm]{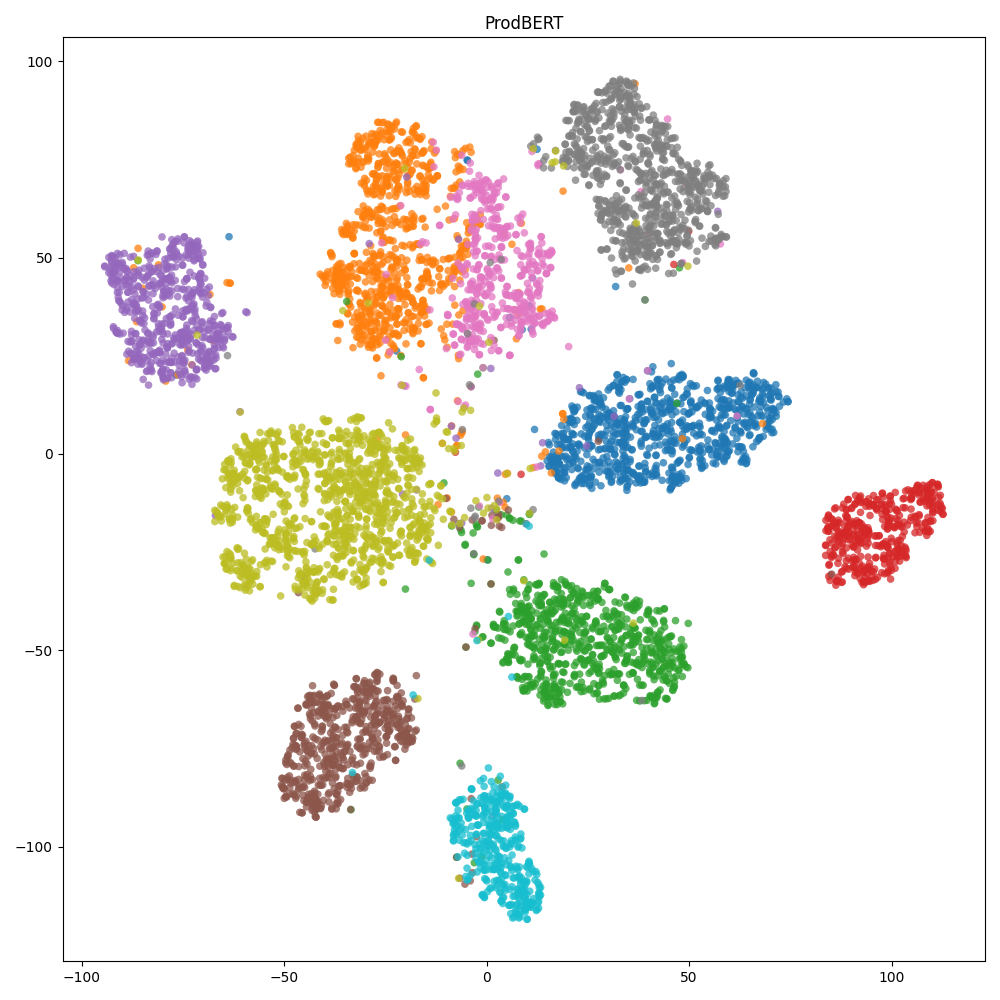} 
    \caption{T-SNE plot of browsing session vector space from Shop B and built with the first hidden layer of pre-trained Prod2BERT.}
    \label{fig:tsne_lm_pb}
\end{figure}

\begin{figure}[!htb]
\centering
    \includegraphics[width=6.3cm]{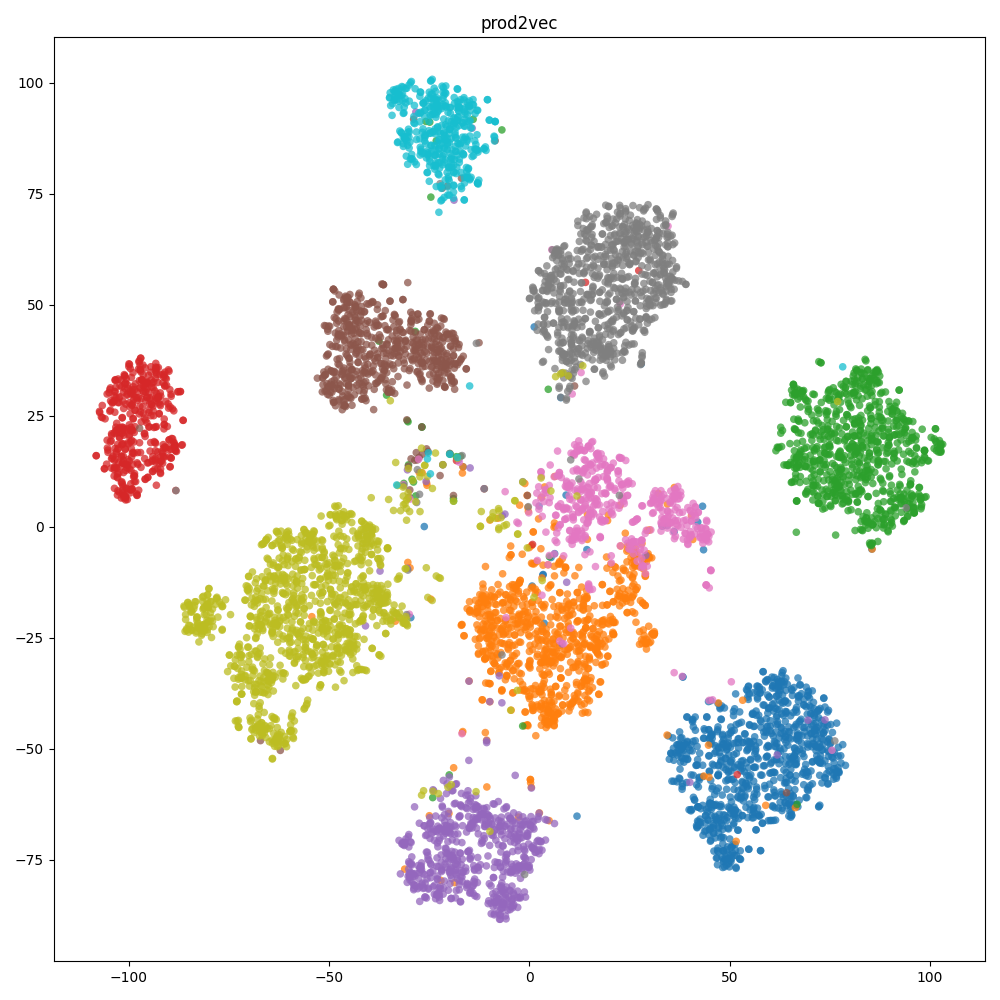} 
    \caption{T-SNE plot of browsing session vector space from Shop B and built with~\textit{prod2vec} embeddings.}
    \label{fig:tsne_lm_p2v}
\end{figure}

Figures~\ref{fig:tsne_ms_pb} to~\ref{fig:tsne_lm_p2v} represent browsing sessions projected in two-dimensions with t-SNE~\cite{tsne}: for each browsing session, we retrieve the corresponding type (e.g. shoes, pants, etc.) of each product in the session, and use majority voting to assign the most frequent product type to the session. Hence, the dots are color-coded by product type and each dot represents a unique session from our logs. It is easy to notice that, first, both contextual and non-contextual embeddings built with a smaller amount of data, i.e. Figures~\ref{fig:tsne_ms_pb} and~\ref{fig:tsne_ms_p2v} from Shop A, have a less clear separation between clusters; moreover, the quality of Prod2BERT seems even lower than~\textit{prod2vec}, as there exists a larger central area where all types are heavily overlapping. Second, comparing Figure~\ref{fig:tsne_lm_pb} with Figure~\ref{fig:tsne_lm_p2v}, both Prod2BERT and~\textit{prod2vec} improve, which confirms Prod2BERT, given enough pre-training data, is able to deliver better separations in terms of product types and more meaningful representations.

\end{document}